\pdfoutput=1

\documentclass[11pt]{article}

\usepackage{amsmath}
\usepackage{multirow}

\usepackage[final]{acl}
\usepackage{cleveref}
\usepackage{times}
\usepackage{latexsym}
\usepackage{booktabs}
\usepackage{amsfonts}
\usepackage{amssymb}
\usepackage{amsthm}
\usepackage{algorithm}
\usepackage{algpseudocode}

\usepackage{subfigure}
\usepackage[T1]{fontenc}

\usepackage[utf8]{inputenc}

\usepackage{microtype}

\usepackage{inconsolata}

\usepackage{graphicx}

\newcommand{\methodname}[0]{{SVIP}}
\title{{\em Draft Model Knows When to Stop:}\\
Self-Verification Speculative Decoding for Long-Form Generation}

\author{
    Ziyin Zhang$^{1,2,}$\thanks{Work done during their internship at Tencent.}\quad
    Jiahao Xu$^{2}$\quad
    Tian Liang$^{2}$\quad
    Xingyu Chen$^{1,2}$\textcolor{darkblue}{\footnotemark[1]}\quad
    Zhiwei He$^{1,2}$\textcolor{darkblue}{\footnotemark[1]}\quad\\
    \textbf{Rui Wang}$^{1}$\thanks{Corresponding authors.} \quad
    \textbf{Zhaopeng Tu}$^{2}$\textcolor{darkblue}{\footnotemark[2]} \\
    $^1$Shanghai Jiao Tong University \qquad \qquad
    $^2$Tencent\\
    $^1$\texttt{\{daenerystargaryen,galaxychen,zwhe.cs,wangrui12\}@sjtu.edu.cn}\\
    $^2$\texttt{\{jettexu,ttianliang,zptu\}@tencent.com}
}

\begin{document}
\maketitle
\begin{abstract}
Conventional speculative decoding (SD) methods utilize a predefined length policy for proposing drafts, which implies the premise that the target model smoothly accepts the proposed draft tokens. However, reality deviates from this assumption: the oracle draft length varies significantly, and the fixed-length policy hardly satisfies such a requirement. Moreover, such discrepancy is further exacerbated in scenarios involving complex reasoning and long-form generation, particularly under test-time scaling for reasoning-specialized models. Through both theoretical and empirical estimation, we establish that the discrepancy between the draft and target models can be approximated by the draft model's prediction entropy: a high entropy indicates a low acceptance rate of draft tokens, and vice versa. Based on this insight, we propose \textbf{\methodname}: \textbf{S}elf-\textbf{V}er\textbf{i}fication Length \textbf{P}olicy for Long-Context Speculative Decoding, which is a training-free dynamic length policy for speculative decoding systems that adaptively determines the lengths of draft sequences by referring to the draft entropy. Experimental results on mainstream SD benchmarks as well as reasoning-heavy benchmarks demonstrate the superior performance of \methodname, achieving up to 17\% speedup on MT-Bench at 8K context compared with fixed draft lengths, and 22\% speedup for QwQ in long-form reasoning.

\end{abstract}

\begin{figure}
    \centering
    \includegraphics[width=0.8\linewidth]{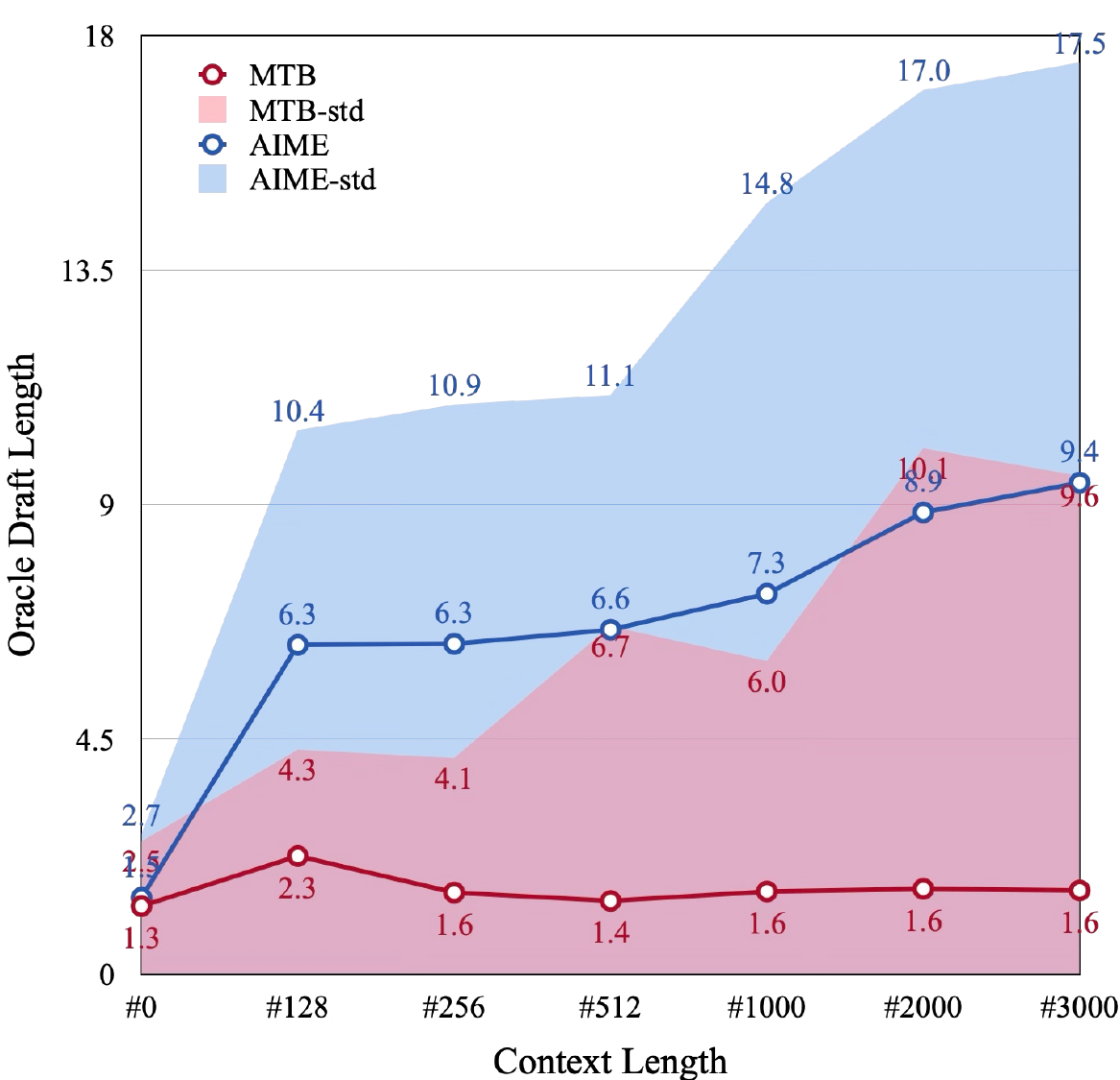}
    \caption{The variance of oracle draft length drastically increases with context length. MTB (MT-Bench): a conventional benchmark for SD systems. AIME: an extremely difficult mathematical testing set for advanced reasoning models.}
    \label{fig:oracle_draft_length}
\end{figure}

\section{Introduction}
Speculative decoding~\citep{2022SD-leviathan,2023SD-chen} is a novel technique that markedly enhances the generation wall-time of large language models (LLMs). This approach employs a small and efficient draft model to draft sequences, while concurrently utilizing a larger and more powerful expert model to verify the drafts. By avoiding the autoregressive generation of each token through the target LLM, speculative decoding achieves improved efficiency while preserving the quality of the output.
This technique is particularly beneficial in the context of {\bf inference-time scaling}, where LLMs generally generate long-form text.

The majority of research on speculative decoding focuses on improving the {\bf acceptance rate of the draft sequences}~\citep{2023SpecTr,2024EAGLE,2024LayerSkip,2024Glide-Cape,2024EAGLE-2,2024MCSD}. However, they limit their settings to a {\bf fixed draft length} (e.g. less than 5 tokens), which we find is sub-optimal in the scenario of long-form text generation. 


\begin{figure*}
    \centering
    \includegraphics[width=\linewidth]{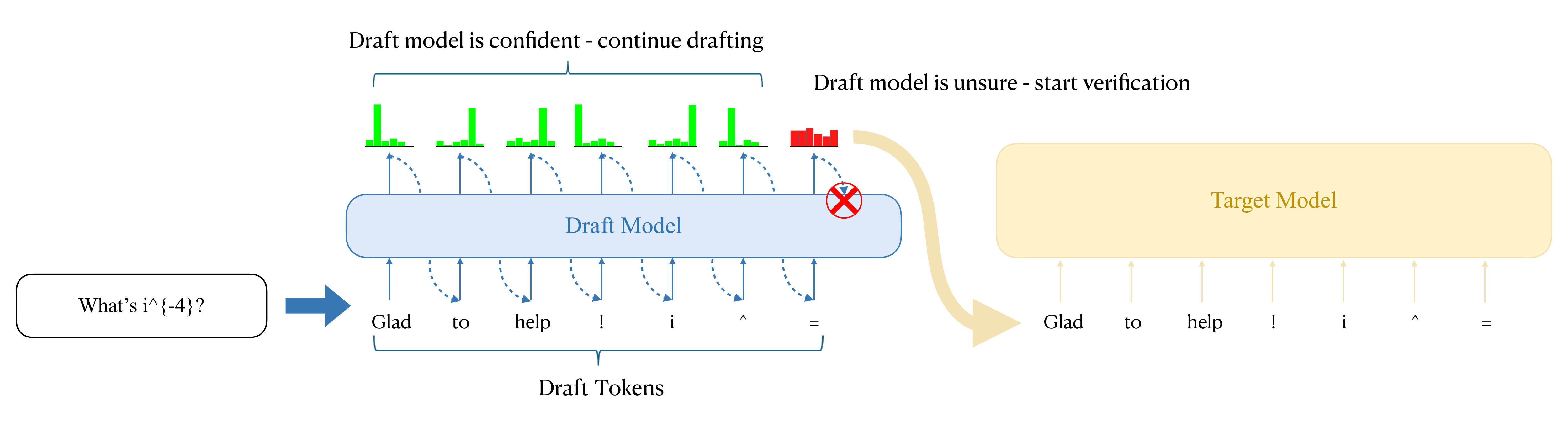}
    \caption{Overview of \textbf{\methodname}: the draft model proceeds the generation process (marked by \textcolor{green}{green}) until it encounters a token for which it has low confidence (marked by \textcolor{red}{red}), signaled by high entropy, at which point the draft model would cease generation and send the draft spans to the target model for verification.}
    \label{fig:overview}
\end{figure*}

Specifically, as discussed in Section~\ref{sec:method-discrepency}, we investigate the token rejection phenomenon in speculative decoding (SD) systems. Our findings reveal that the oracle draft length varies considerably for long-form generation, as shown in \cref{fig:oracle_draft_length}. Furthermore, heuristic methods, such as predicting the draft length, are impractically challenging because our further investigation reveals that token rejection occurs unexpectedly. Moreover, our closer investigation indicates a strong correlation between token rejection and the draft model's prediction entropy at that moment.

Inspired by such a correlation, we analyze the acceptance rate and propose our \textbf{\methodname}: \textbf{S}elf-\textbf{V}er\textbf{i}fication Length \textbf{P}olicy in Section~\ref{sec:method-lower-bound}. Specifically, we derive a lower bound for the acceptance rate based on the entropy information from the draft model. Notably, \methodname~not only approximates this lower bound but also dynamically adjusts the length of draft sequences by determining whether to continue drafting or initiate verification after each token generation, as shown in Figure~\ref{fig:overview}. By optimizing draft sequence lengths, \methodname~enhances SD systems' efficiency. Importantly, our method is entirely training-free and thus can be seamlessly integrated with any SD decoding algorithm, making it broadly applicable and efficient.

With extensive experiments across multiple model sizes and evaluation benchmarks, we demonstrate the superior performance of \methodname~ in long-context generation. Compared with fixed-length draft policies, it yields up to 17\% speedup on MT-Bench~\citep{2023MT-Bench} and 22\% on AIME. As a training-free length policy, \methodname\ is also extremely flexible and compatible with state-of-the-art speculative decoding systems such as EAGLE-2~\citep{2024EAGLE-2}, achieving an additional 13\% speed improvement.


In summary, our contributions are threefold:
\begin{enumerate}
    \item We provide an in-depth analysis of the disagreement between draft model and target model in speculative decoding systems, highlighting the underperformance of fixed-length draft length policies.
    \vspace{-0.1cm}
    \item Based on this analysis, we derive a low bound of speculative decoding systems, where the acceptance rate of the draft model could be modeled by its entropy only. We further develop \methodname, an entropy-based dynamic draft length policy for speculative decoding systems, which is extremely flexible and can be adapted to any auto-regressive draft model.
    \vspace{-0.1cm}
    \item Experimental results demonstrate the superior performance of \methodname\ over baseline draft length policies on both conventional long-form generation and reasoning-heavy benchmarks.
\end{enumerate}

\section{Draft Model Knows When to Stop}\label{sec:method}
In this section, we first examine the behavior of draft models at the rejection phenomenon, and analyze the oracle lengths for SD systems. Then, we theoretically derive \methodname, which approximates the draft token acceptance rate using the draft model's own prediction entropy.


\subsection{Investigation of Rejection}
\label{sec:method-discrepency}


Speculative decoding enhances the efficiency of large language model (LLM) inference by assuming that draft tokens are \emph{accepted} by the LLM, thus avoiding autoregressive generation. 
Should the target model exhibit a tendency to reject tokens, the overall performance of the system may experience considerable degradation.
Consequently, empirically investigating the rejection phenomenon is our primary interest.

Specifically, we analyze the distribution characteristics of rejected tokens across two scenarios: 
\begin{itemize}
    \setlength{\itemsep}{0pt}
    \item \textbf{AIME (2022-2024)}: a challenging math reasoning dataset, using greedy decoding with QwQ-32B-Preview~\citep{2024QwQ} and a 1.5B draft model;
    \item \textbf{MT-Bench}: a conversational and instruction-following benchmark, using sampling decoding with the Qwen2.5 family~\citep{2024Qwen2}.
\end{itemize}

\begin{figure}
    \centering
    \subfigure[KL Div. ($\downarrow$)]{\includegraphics[width=0.5\linewidth]{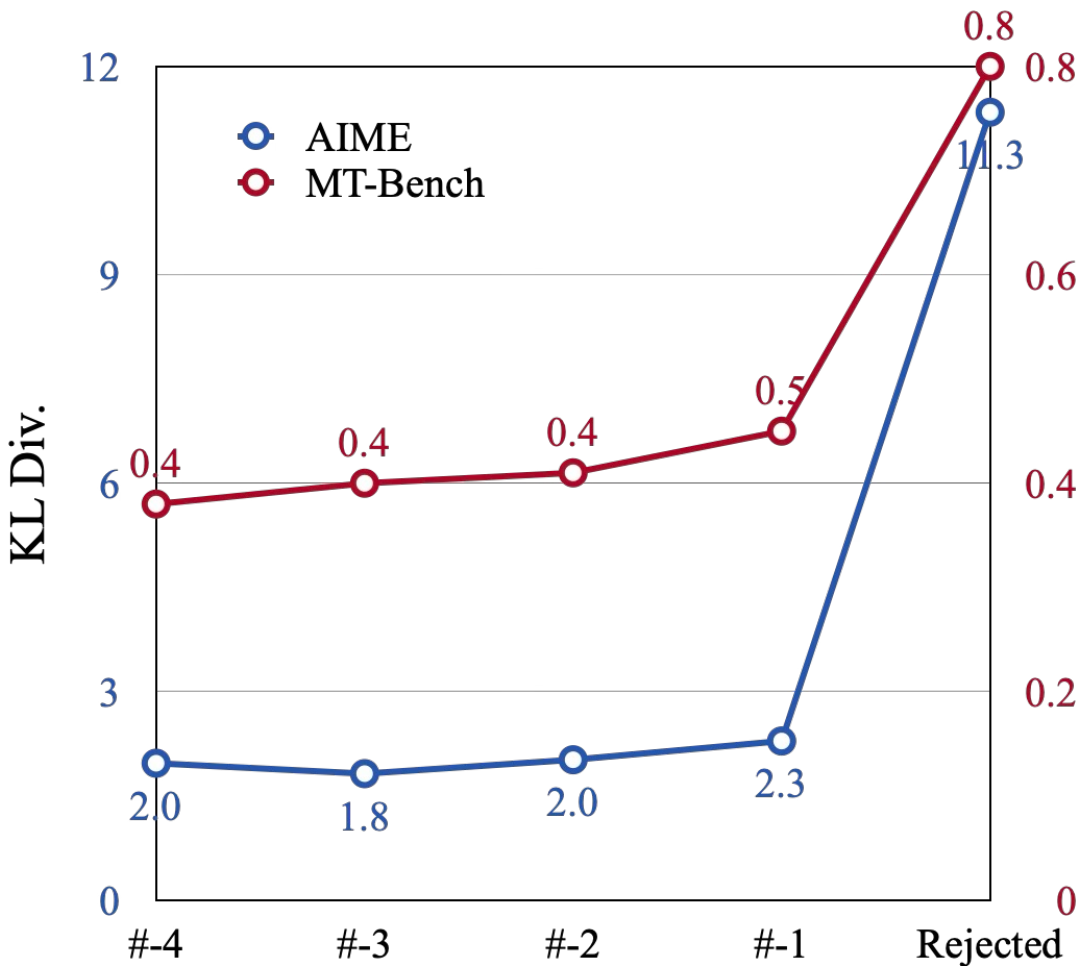}\label{fig:kl}}
    \subfigure[Sorted vocab. distribution]{\includegraphics[width=0.48\linewidth]{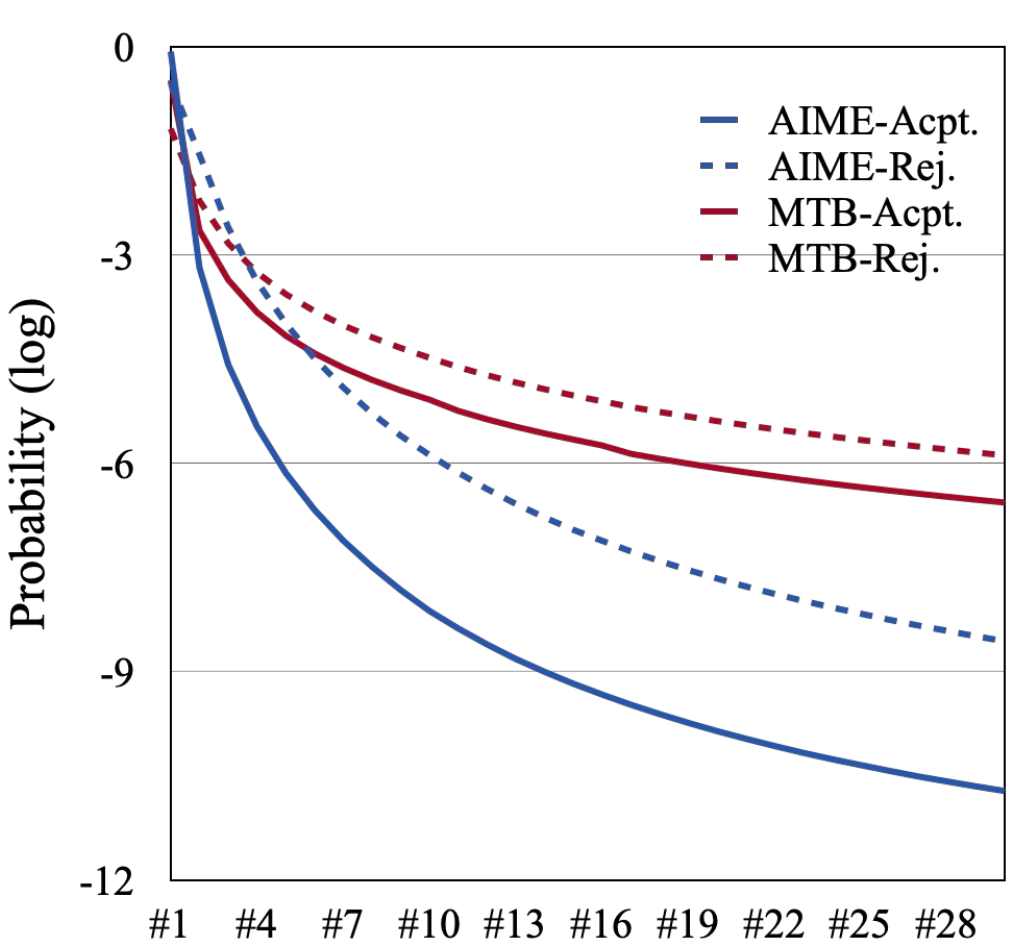}\label{fig:vocab_logp}}
    \caption{Agreement scores and sorted vocabulary log probability at the rejection phenomenon. The x-axis of the \Cref{fig:kl} represents rejected tokens and the four tokens before them.}
\end{figure}


\paragraph{Rejection occurs out of the blue}
A natural question is: \textit{How does the token rejection occur? Is there any symptom before it happens?} We investigate the rejected tokens with the following metrics:
\begin{itemize}
    \setlength{\itemsep}{0pt}
    \item KL divergence for vocabulary distribution difference. 
    \item Vocabulary distribution of accepted and rejected tokens.
\end{itemize}
\noindent We quantify the occurrence of the token rejection phenomenon along with its corresponding prefix tokens, as illustrated in \Cref{fig:kl,fig:vocab_logp}. It shows that KL metrics experienced a sudden and substantial surge in the position of the rejected token, and the output vocabulary distribution at rejected tokens differs significantly from previously correctly drafted tokens. Such a phenomenon is expected: KL divergence indicates a system discrepancy with rejection, where the ground-truth tokens are inherently difficult for the draft model to predict. We also quantify the entropy of the draft model at the acceptance and rejection positions in Table~\ref{tab:entropy}. It shows that the draft models suffer from a severely high entropy at rejection, indicating extreme difficulty in modeling the corresponding tokens.

\begin{table}[htbp]
\centering

\resizebox{0.6\linewidth}{!}{%
    \begin{tabular}{lrr} 
    \toprule
    \multirow{2}{*}{\textbf{Dataset}} & \multicolumn{2}{c}{\textbf{Vocab Entropy}} \\ 
    \cmidrule{2-3}
     & \textit{Accepted.} & \textit{Rejected} \\ 
    \midrule
    AIME & 0.25 & 1.30 \\
    MT-Bench & 2.18 & 3.99 \\
    \bottomrule
    \end{tabular}
}
\caption{Vocabulary entropy of the draft model at accepted and rejected draft tokens.}
\label{tab:entropy}
\end{table}


\subsection{Our Method: SVIP}
Since the rejection phenomenon occurs suddenly and the draft model suffers from high entropy at the rejection position, can we detect the rejection with KL divergence or draft model's entropy? To achieve this,
we seek a theoretical understanding of the rejection phenomenon.

\paragraph{Lower Bound of Acceptance}
Since rejection denotes the sudden decrease of acceptance rate, we investigate the theoretic acceptance rate of the SD systems. Specifically, given a target model $p$, a draft model $q$, an input sequence $x_{< t}$, and a draft token $x_t$, it's easy to derive that $x_t$'s acceptance probability is $\min\left(1, \; \frac{p(x_t)}{q(x_t)}\right)$ (see Appendix~\ref{appendix:sd-algorithm}). Let $\beta$ denote the expected acceptance probability over the distribution of $x_t$, and it follows that:
\begin{align}
    \beta   & = \sum_xq(x)\cdot\min\left(1, \frac{p(x)}{q(x)}\right)\nonumber\\
            & = \sum_x \min\left(p(x), q(x)\right), \label{eq:beta}
\end{align}
\noindent where $p$ and $q$ denote the target and draft model respectively. \citet{2023SD-chen} has proven that $\beta$ is related to the total variational distance (TVD) between $p$ and $q$. Start from this, we utilize Pinsker's inequality in \Cref{eq:pinsker} and yield the following bound in \Cref{eq:beta-bound}:
\begin{align}
    \beta &= 1 - \mathrm{TVD}(p, q) \label{eq:tvd} \\
    &\geqslant 1 - \sqrt{\frac{1}{2}\mathbb{KL}(q||p)} \label{eq:pinsker} \\
    &= 1 - \sqrt{\frac{1}{2}H_{q,p} - \frac{1}{2}H_q} \label{eq:beta-bound}
\end{align}
\noindent where $H_{q,p}$ is the cross entropy between $q$ and $p$, and $H_q$ is the entropy of $q$. 
We denote the above bound as the \textit{oracle bound}. 
Utilizing this bound for acceptance prediction is infeasible since it requires instantaneous access to the target model for cross entropy $H_{q,p}$, which is infeasible during the drafting phase. 

\paragraph{Approximating the Oracle Bound with Draft Distribution}
Can we approximate the cross-entropy $H_{q,p}$ between the draft model’s distribution $q$ and the target distribution $p$ using only $q$? We propose using the draft model’s entropy $H_q$ as a proxy, approximating $H_{q,p}$ as $\gamma H_q$, where $\gamma = H_{q,p} / H_q$ is a random variable capturing the ratio between $H_{q,p}$ and $H_q$. This leads to a bound on the acceptance rate $\beta$:
\begin{equation}
    \beta \geq 1 - \sqrt{\frac{1}{2} (\gamma - 1) H_q}.
\end{equation}
To make this bound practical, we approximate $\gamma$ with a constant $c$, yielding the \textit{approximation bound}:
\begin{equation}
    \beta \geq 1 - \sqrt{\frac{1}{2} (\gamma - 1) H_q} \approx 1 - \sqrt{c H_q}. \label{eq:appro_bound}
\end{equation}
This bound holds when $1 - \sqrt{\frac{1}{2} (\gamma - 1) H_q} \geq 1 - \sqrt{c H_q}$, ensuring the approximation is conservative. Thus, the approximation bound $1 - \sqrt{c H_q}$ lower-bounds the true acceptance rate $\beta$.

\paragraph{Detecting Rejection with Draft Entropy}\label{sec:method-lower-bound}

With \Cref{eq:appro_bound} providing a way to estimate the acceptance probability using only the draft model’s entropy, we introduce \textsc{\methodname}, which dynamically adapts the draft length. After generating each draft token, we compute the approximation bound and halt drafting if it falls below a threshold $\hat{h}$, i.e., if $1 - \sqrt{c H_q} < \hat{h}$. Since $c$ and $\hat{h}$ are constant hyperparameters, we simplify the criterion by defining a new threshold $h = (1 - \hat{h}) / \sqrt{c}$, absorbing $\sqrt{c}$ into $h$. This reduces the stopping condition to $\sqrt{H_q} > h$. Formally, given a prefix $x_{<t}$ of $t-1$ tokens, the stopping criterion is:
\begin{equation}
    \sqrt{H_q(x_{<t})} > h, \label{eq:stop_condition}
\end{equation}
where $H_q(x_{<t})$ is the entropy of the draft distribution conditioned on $x_{<t}$. This ensures drafting stops when the estimated acceptance probability is too low, optimizing efficiency while maintaining reliability. We formalize \methodname\ in Algorithm~\ref{algo:sd-dynamic}. The details of the methods \texttt{Verify} and \texttt{Correct} are given in Appendix~\ref{appendix:sd-algorithm}, for which different versions are available for sampling (Algorithm~\ref{algo:verify-sample}, \ref{algo:correct-sample}) and greedy decoding (Algorithm~\ref{algo:verify-greedy}, \ref{algo:correct-greedy}).

\begin{algorithm}[t]\small
\caption{\methodname}
\label{algo:sd-dynamic}
    \begin{algorithmic}[1]
        \Require target model $p$, draft model $q$, input sequence $x_{\leqslant t}$, maximum length $T$, threshold $h$
        \vspace{3pt}
        \State Initialize $n\gets t$
        \While{$n<T$}
            \State $j=0$
            \While{True}
                \State Sample $x_{n+j} \sim q(x|x_{<{n+j}})$
                \State $j\gets j+1$
                \If{$\sqrt{H(q_{x|x_{<n+j}})} > h$}
                    \State Exit while loop
                \EndIf
            \EndWhile
            \State $\gamma\gets j$
            \State Compute $p(x|x_{<{n+j}}),\; j=1, \cdots, \gamma+1$ in parallel
            \State $\tilde n \gets n$
            \For{$j=1$ to $\gamma$}
                \If{Verify$\left(p_{x|x_{<n+j}},\, q_{x|x_{<n+j}}, x_{n+j}\right)$}
                    \State $\tilde n \gets \tilde n + 1$
                \Else
                    \State $x_{n+j} \gets \text{Correct}\left(p_{x|x_{<n+j}},\, q_{x|x_{<n+j}}\right)$
                    \State Exit for loop
                \EndIf
            \EndFor
            \If{$\tilde n == n+\gamma$}
                \State Sample $x_{n+\gamma+1}$ from $p(x|x_{\leqslant n+\gamma})$
            \EndIf
            \State $n\gets\tilde n + 1$
        \EndWhile
        \vspace{3pt}
        \Ensure $x_{\leqslant n}$
    \end{algorithmic}
\end{algorithm}

\begin{figure}[htbp]
    \centering
    \includegraphics[width=\linewidth]{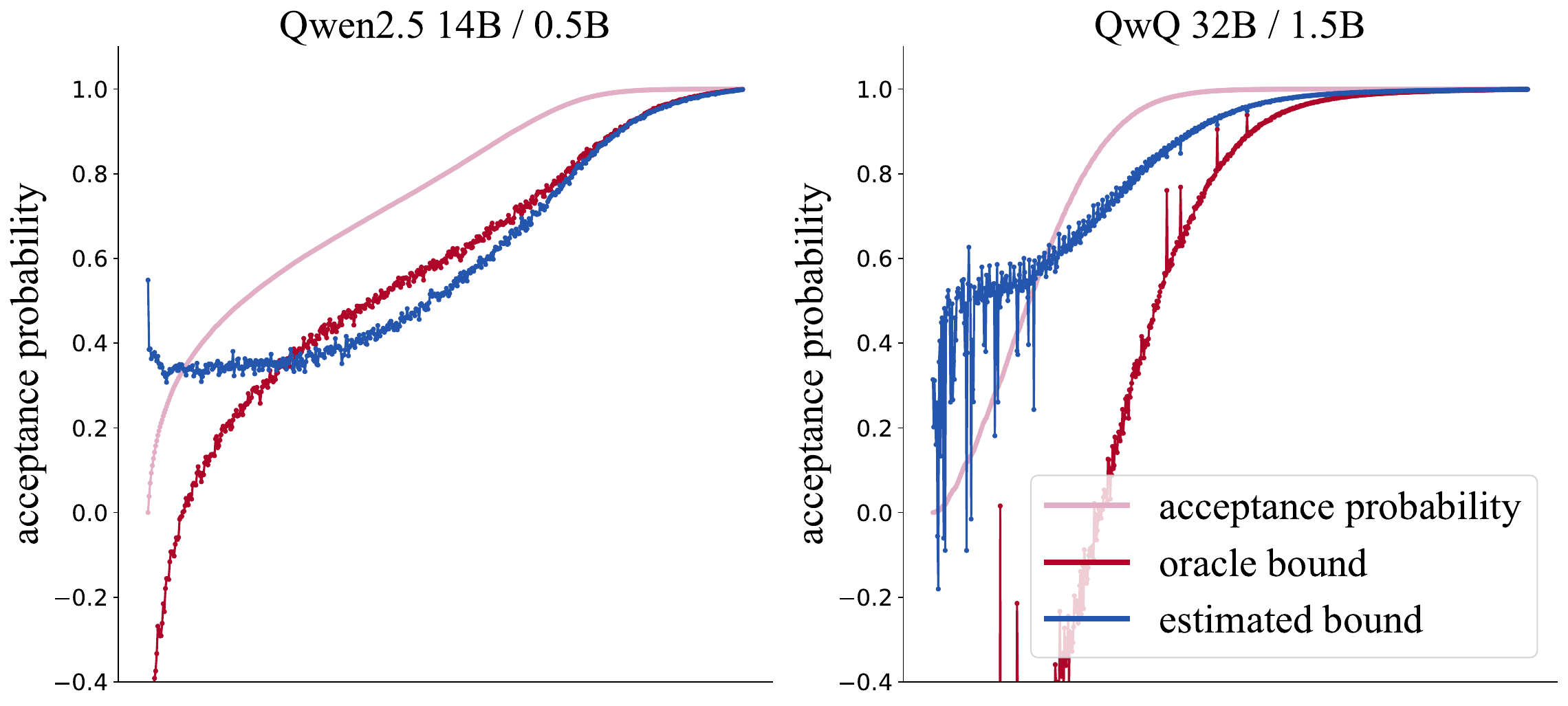}
    \caption{Comparison between the actual acceptance probability in \Cref{eq:beta}, the acceptance probability lower bound in \Cref{eq:beta-bound}, and the estimated lower bound in \Cref{eq:appro_bound}. Each position on the x-axis corresponds to a token, which has been sorted according to the actual acceptance probability.}
    \label{fig:acceptance-prob-estimation}
\end{figure}

\begin{figure*}[ht]
    \centering
    \includegraphics[width=\linewidth]{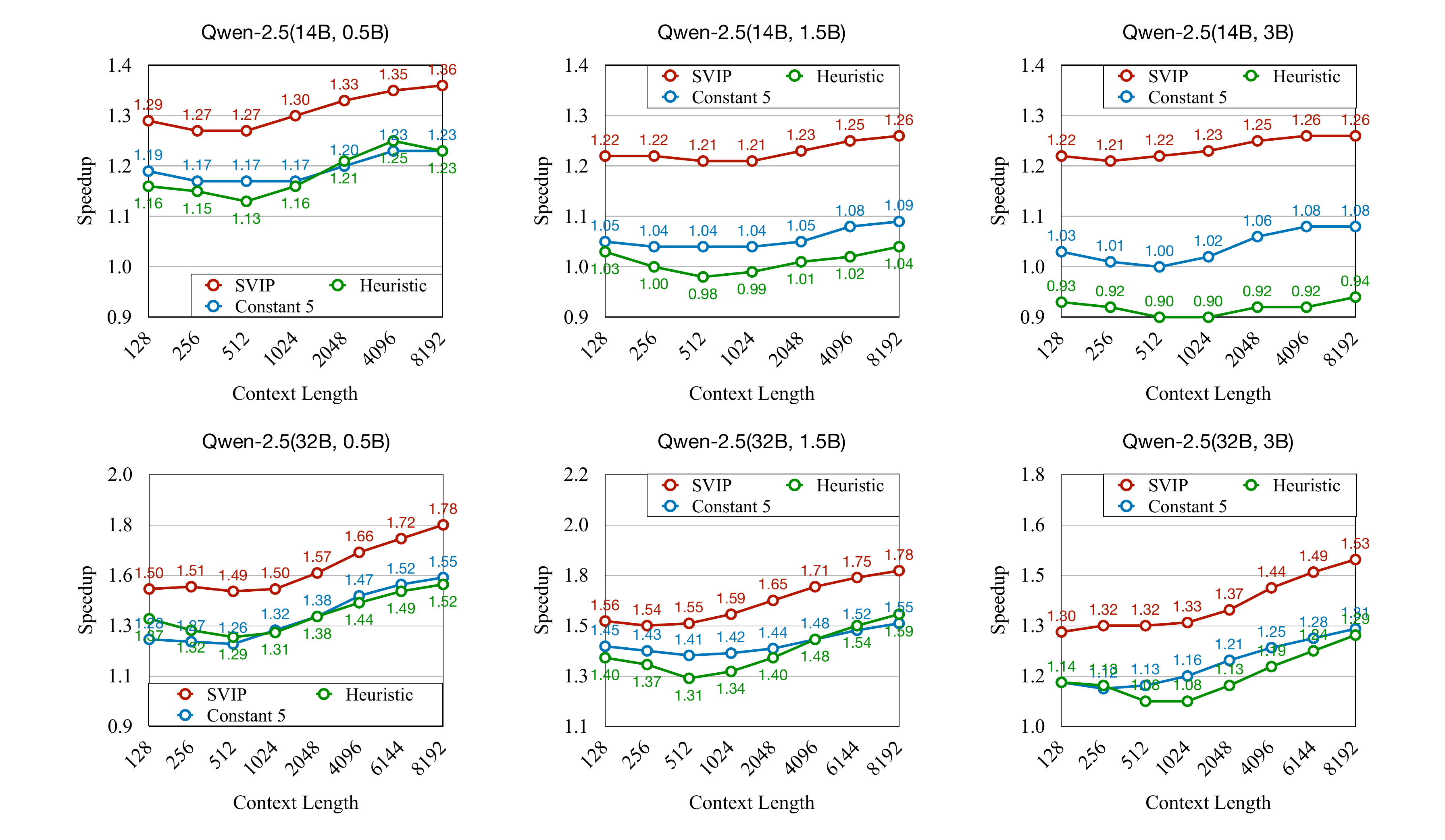}
    \caption{The SD system speedup on MT-Bench using Qwen2.5-14B (top) and Qwen2.5-32B (bottom) as targets and three different smaller models as drafts.}
    \label{fig:results_qwen14b}
\end{figure*}

\paragraph{Justifying the Approximation Bound}
The tightness of the approximation depends on how well $c$ captures the behavior of the random variable $\gamma$. For the approximation bound to be conservative, it requires:
\begin{equation}
    \gamma \leq 2c + 1.
\end{equation}
Given the right-skewed nature of $\gamma \geq 1$, we model $\gamma = 1 + X$, where $X \sim \text{Gamma}(\alpha, \beta)$. The probability that the approximation is valid is:
\begin{equation}
    P(\gamma \leq 2c + 1) = P(X \leq 2c) = \frac{\gamma(\alpha, \beta \cdot 2c)}{\Gamma(\alpha)}, \label{eq:gamma_assump}
\end{equation}
where $\gamma(\alpha, z) = \int_0^z t^{\alpha-1} e^{-t} \, dt$ is the lower incomplete gamma function, and $\Gamma(\alpha)$ is the gamma function. The choice of $c$ trades off reliability and tightness:
\begin{itemize}
    \item If $c$ is small (e.g., $2c < \mathbb{E}[X] = \alpha / \beta$), the probability $P(X \leq 2c)$ is low, risking an invalid bound.
    \item If $c$ is large, $P(X \leq 2c) \to 1$, but the bound $\beta \geq 1 - \sqrt{c H_q}$ becomes looser.
\end{itemize}
Optimal performance requires balancing the reliability of the approximation (high $P(X \leq 2c)$) with the tightness of the bound (small $c$).

We analyze such a trade-off on Qwen2.5 on MT-Bench and QwQ-32B on AIME in \Cref{fig:acceptance-prob-estimation}. It shows that for most cases our estimated approximation bound works well (has a higher acceptance probability than the oracle bound when hyperparameter $c$ (i.e. $h$) is properly selected, set to 0.18 in the figure).



\section{Experiments}\label{sec:experiments}

Next, to verify the effectiveness of \methodname, we conduct experiments on both conventional long-form generation~(Section~\ref{sec:experiments-mtbench}) and reasoning with test-time scaling~(Section~\ref{sec:experiments-qwq}). Since \methodname\ is completely training-free, we also apply it to other speculative decoding methods and demonstrate its flexibility~(Section~\ref{sec:experiments-eagle}).

As baselines, we consider two widely adopted policies for draft length: 
\begin{enumerate}
\setlength{\itemsep}{0pt}
    \item \textbf{Constant}: a constant draft length (set to 5 unless otherwise stated), which is commonly used in the literature
    \item \textbf{Heuristic}: the heuristics implemented in Hugging Face Transformers library~\citep{2019transformers}, where the draft length for the next draft iteration is increased by 2 if all draft tokens in the current iteration are accepted, and otherwise decreased by 1.
\end{enumerate}

\subsection{Results on Long-form Generation}
\label{sec:experiments-mtbench}
\paragraph{Settings}
We first validate the effectiveness of \methodname\ on the widely used MT-Bench~\citep{2023MT-Bench} using different sizes of Qwen2.5~\citep{2024Qwen2} as target and draft models. Unlike many existing works on speculative decoding~\citep{2023SD-chen,2024Glide-Cape} that limit their experiments to generating short sequences of 128 tokens, we conduct experiments on long-form generation with up to 8K context to investigate the applicability of speculative decoding in a broader scope. We set the sampling temperature to 1, as we found that when using greedy decoding in long-form generation, both the draft and the target models are prone to repeat themselves, resulting in very low information entropy and exaggerated speedup ratios~\citep{2024temperature-investigation} (see Appendix~\ref{appendix:longform-results} for details). The entropy threshold $h$ in \methodname\ is chosen from $\{0.2, 0.3, 0.4, 0.5\}$ based on performance on 8 held out samples using the 14B model as target and the 0.5B model as draft, which is set to 0.3 and reused in all following experiments.

As evaluation metrics, we mainly report the average speedup over target-model-only autoregressive decoding, but also consider other auxiliary information including accepted draft lengths and draft token accept rate. Also, since the memory consumption of verifying $n$ draft tokens is quadratic in $n$, we limit the maximum draft length to 40 in both heuristics and \methodname\ scenarios, beyond which we start to encounter out-of-memory issues.

\paragraph{\methodname~outperforms all baselines} We validate our proposed method, \methodname, with two target models: Qwen2.5-14B and Qwen2.5-32B, utilizing draft models that vary in size from 0.5B to 3B.
We show the performance of \methodname\ and baselines in Figure~\ref{fig:results_qwen14b}. It shows that \methodname\ consistently outperforms constant and heuristics draft length by a large margin.

\begin{figure}[t]
    \centering
    \includegraphics[width=1\linewidth]{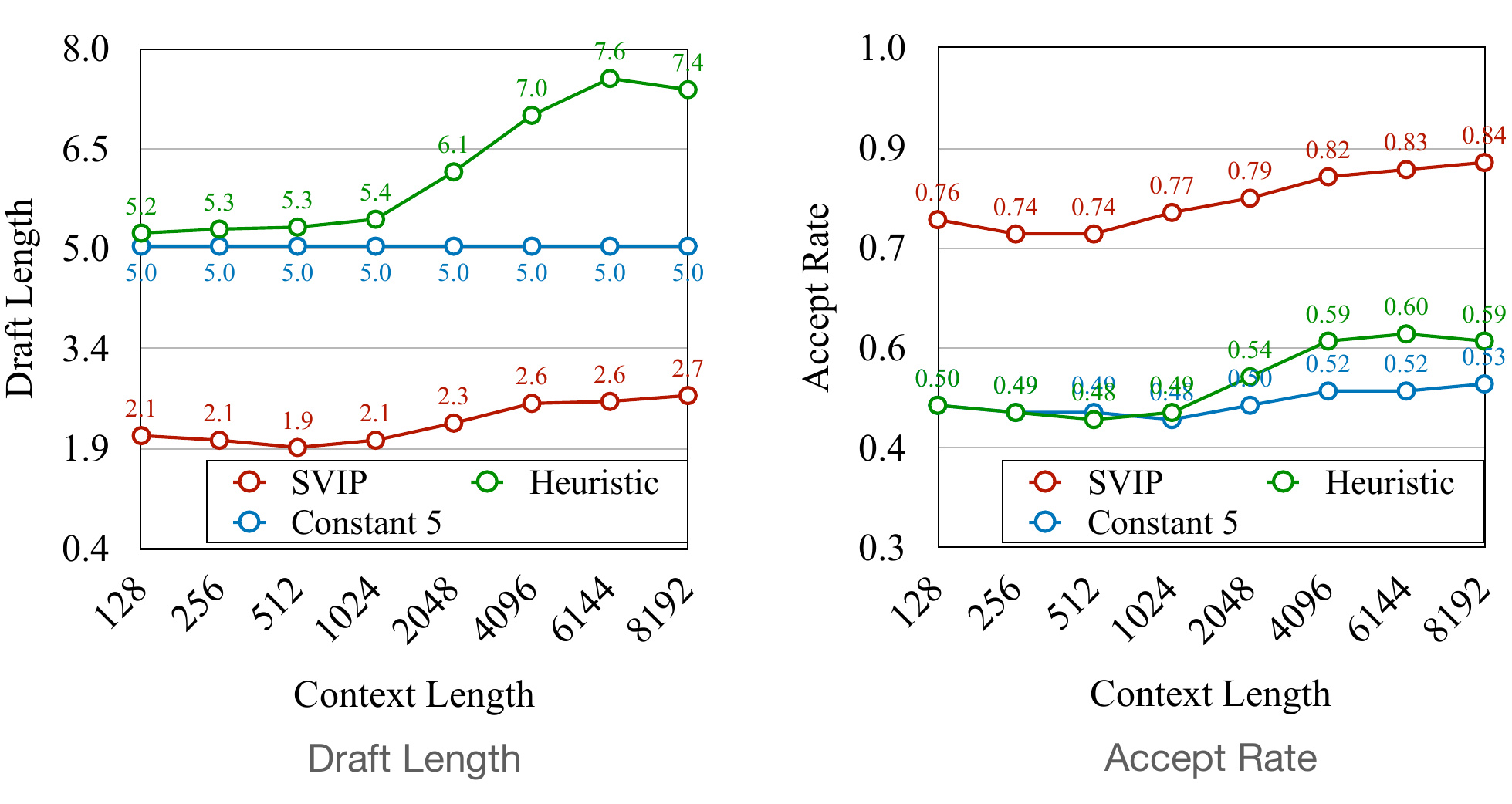}
    \caption{Analysis of Qwen2.5 14B/0.5B's behaviours on MT-Bench. Compared with constant and heuristics length policies, \methodname\ generates shorter drafts with a significantly higher accept rate.}
    \label{fig:results_draft_len}
\end{figure}
\begin{figure}[t]
    \centering
    \vspace{-0.3cm}
    \subfigure[]{\includegraphics[width=0.48\linewidth]{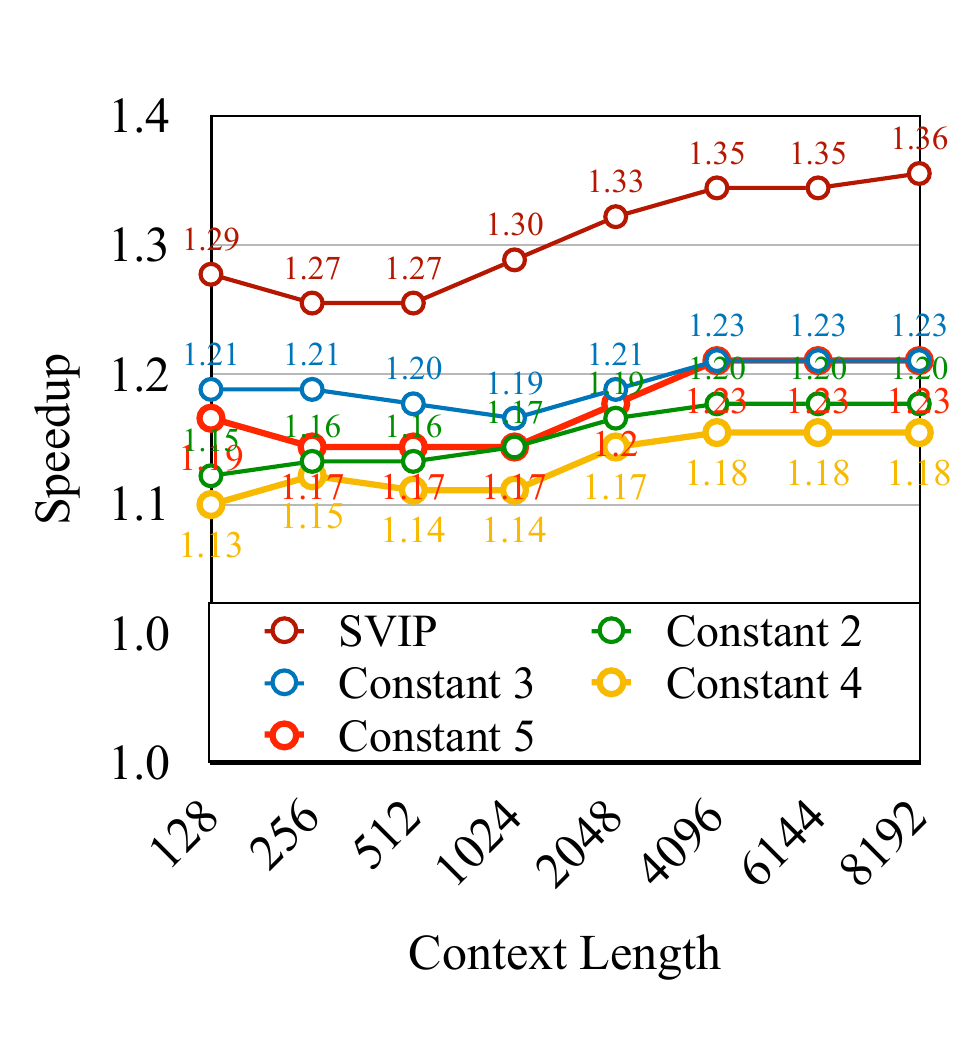}\label{fig:results_const2-5}}
    \subfigure[]{\includegraphics[width=0.48\linewidth]{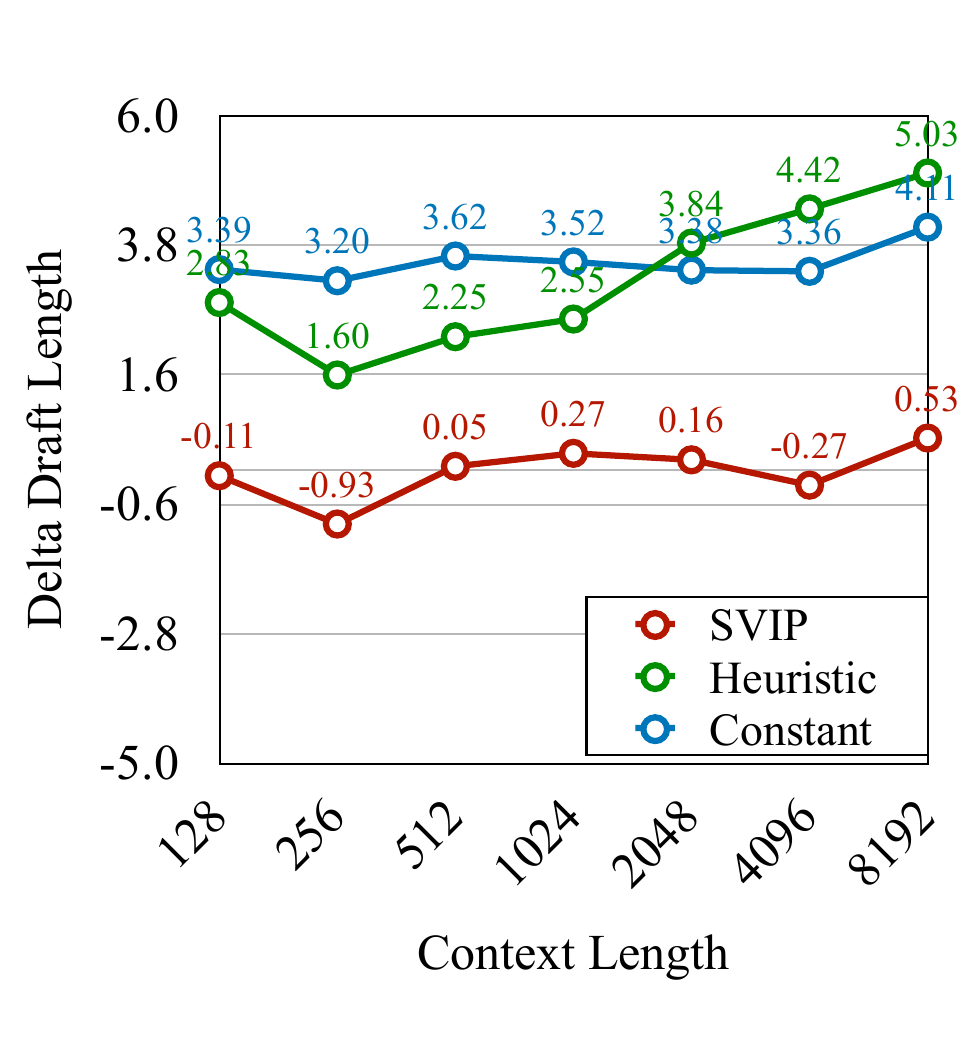}\label{fig:delta-draft-len}}
    \caption{Further analysis of Qwen2.5 14B/0.5B on MT-Bench: (a) \methodname\ outperforms all constant length policies ranging from 2-5; (b) delta draft length (defined as proposed draft length minus oracle draft length) show that constant and heuristics length policies tend to over-generate drafts, while \methodname\ models the oracle draft length almost perfectly.}
    \vspace{-0.5cm}
\end{figure}

To further investigate the origin of \methodname's speedup, we analyze the proposed draft lengths and accepte rate of the different length policies in Figure~\ref{fig:results_draft_len}. We observe that by terminating the draft process early when the draft model entropy is high, \methodname\ leads to shorter draft lengths and a much higher acceptance rate. However, in Figure~\ref{fig:results_const2-5} we also compare the performance of \methodname\ with shorter constant draft length policies, which suggest that simply using shorter constant draft length does not suffice, highlighting the importance of dynamically determining draft lengths. Figure~\ref{fig:results_draft_len} also suggests that while the heuristics draft length policy tends to produce very long drafts at long context, its acceptance rate remains low, resulting in no effective speedup compared with fixed draft length.

\paragraph{\methodname~ drafts fit oracle length} In Figure~\ref{fig:delta-draft-len}, we revisit the concept of ``oracle draft length'' introduced in Section~\ref{sec:method-discrepency}, and plot the differences between actual draft lengths and oracle draft lengths. The results suggest that both constant and heuristics draft length policies tend to generate drafts that are too long, while \methodname\ models the oracle draft length almost perfectly, with an average delta below 0.5 tokens.


\paragraph{\methodname~further boosts strong baseline}\label{sec:experiments-eagle}
In the previous experiments, we evaluated \methodname\ on vanilla speculative decoding, where a standard pretrained Transformer decoder model with the same vocabulary as the target is used as the draft model. However, in the past years many works on speculative decoding have proposed other stronger or more efficient draft models~\citep{2024Medusa,2024Glide-Cape,2024EAGLE}. Since most of these works assume a constant draft length, \methodname\ is orthogonal to them and can be applied on top of them without any additional training.

\begin{table}[t]
\centering
\caption{Speedup on MT-Bench on top of EAGLE-2, using Vicuna as base models.}
\label{tab:eagle}
\resizebox{1\linewidth}{!}{%
\begin{tabular}{llcccccc} 
    \toprule
        \multirow{2}{*}{\textbf{Model}} & \multirow{2}{*}{\textbf{Method~}}       & \multicolumn{6}{c}{\textbf{Context Length}}    \\ 
    \cmidrule{3-8}
    \multicolumn{2}{l}{}     & 128  & 256  & 512  & 1K   & 2K   & 4K\\ 
    \midrule
        \multirow{2}{*}{V-7B}   & E2   & 2.70 & 2.62 & 2.52 & 2.41 & 2.43 & 1.24 \\
       & \, + \methodname & \textbf{2.80} & \textbf{2.76} & \textbf{2.71} & \textbf{2.69} & \textbf{2.75} & \textbf{1.41} \\ 
    \midrule
        \multirow{2}{*}{V-13B}  & E2   & \textbf{2.95} & 2.90 & 2.83 & 2.74 & 2.71 & 1.53 \\
       & \, + \methodname & 2.94  & \textbf{2.99}  & \textbf{2.93}  & \textbf{2.86}  & \textbf{2.79}  & \textbf{1.64} \\
    \bottomrule
\end{tabular}
}

\end{table}

Specifically, we also apply \methodname\ to EAGLE-2~\citep{2024EAGLE-2}, the current state-of-the-art (SOTA) speculative decoding system which utilizes the target model's language modeling head on top of the draft model's features to predict the next draft token, and dynamically constructs a draft tree at each draft position. Following \citet{2024EAGLE-2}, we use Vicuna 7B, 13B~\citep{2023Vicuna} as the base models, and set the sampling temperature to 1. To the best of our knowledge, we are also the first to investigate EAGLE-2's effectiveness in long-form generation.
The results of EAGLE-2 on MT-Bench are given in Table~\ref{tab:eagle}. Even on top of this SOTA speculative decoding system, \methodname\ yields consistent improvement, which is especially notable at longer context length, surpassing the vanilla EAGLE-2 by 14\% speedup for Vicuna 7B and 7\% for Vicuna 13B.




\begin{table*}
\centering
\caption{Speedup of QwQ on MATH, AIME, and GPQA, along with their average generation length.}
\label{tab:qwq}
\begin{tabular}{lcccccccc} 
\toprule
    & \multicolumn{5}{c}{\textbf{MATH500}} & \multirow{2}{*}{\textbf{GPQA}} & \multirow{2}{*}{\textbf{AIME}} & \multirow{2}{*}{\textbf{Avg}} \\ 
\cmidrule{2-6}
    & Level1  & Level2  & Level3  & Level4   & Level5     \\ 
\cmidrule{2-9}
    Avg. Length & 1.3K & 1.3K & 1.6K & 2.5K & 3.6K & 3.9K  & 6.2K\\
\midrule
    Const. & 1.45 & 1.50 & 1.52 & 1.56 & 1.56 & 1.25 & 1.58 & 1.49 \\
    Heuristics & 1.29 & 1.26 & 1.27 & 1.30 & 1.33 & 1.18 & 1.34 & 1.28 \\
    \methodname & \textbf{1.65} & \textbf{1.68} & \textbf{1.75} & \textbf{1.78} & \textbf{1.82} & \textbf{1.52} & \textbf{1.77} & \textbf{1.71}\\
\bottomrule
\end{tabular}

\end{table*}


\subsection{Long-form Reasoning}\label{sec:experiments-qwq}
\paragraph{Settings}
Recently, o1-style reasoning models have come into the spotlight of LLM research. Thus, we are especially interested in seeing the effectiveness of SVIP and other speculative decoding strategies on such models, which often have very long outputs. Consequently, we utilize QwQ-32B-Preview~\citep{2024QwQ}, the only applicable open-source reasoning model at the time of writing, which does not have off-the-shelf smaller variants, so we train our own draft model based on Qwen2.5 1.5B by distilling QwQ 32B on 1M mathematical Persona data~\citep{2024Persona}\footnote{We have made this model publicly available and will provide links in camera-ready version.}. Using this draft model, we conduct experiments on MATH~\citep{2021MATH}, AIME\footnote{\url{https://maa.org/maa-invitational-competitions/}}, and GPQA~\citep{2023GPQA}. We sample 200 questions from MATH ranging from level 1 to level 5, and use 73 questions released from 2022 to 2024 for AIME. For GPQA, we use the diamond test set. 

\paragraph{\methodname~achieves strong speedup in long-form reasoning}
The overall results are given in Table~\ref{tab:qwq}. \methodname\ outperforms the two baselines by a large margin across different benchmarks and context lengths. Detailed analysis of the different length policies' behaviours suggests similar results to the previous experiment: \methodname\ has an average proposal length similar to the constant draft length policy, but with a much higher draft token accept rate, leading to more effective speedup.


\begin{figure}
    \centering
    \includegraphics[width=0.8\linewidth]{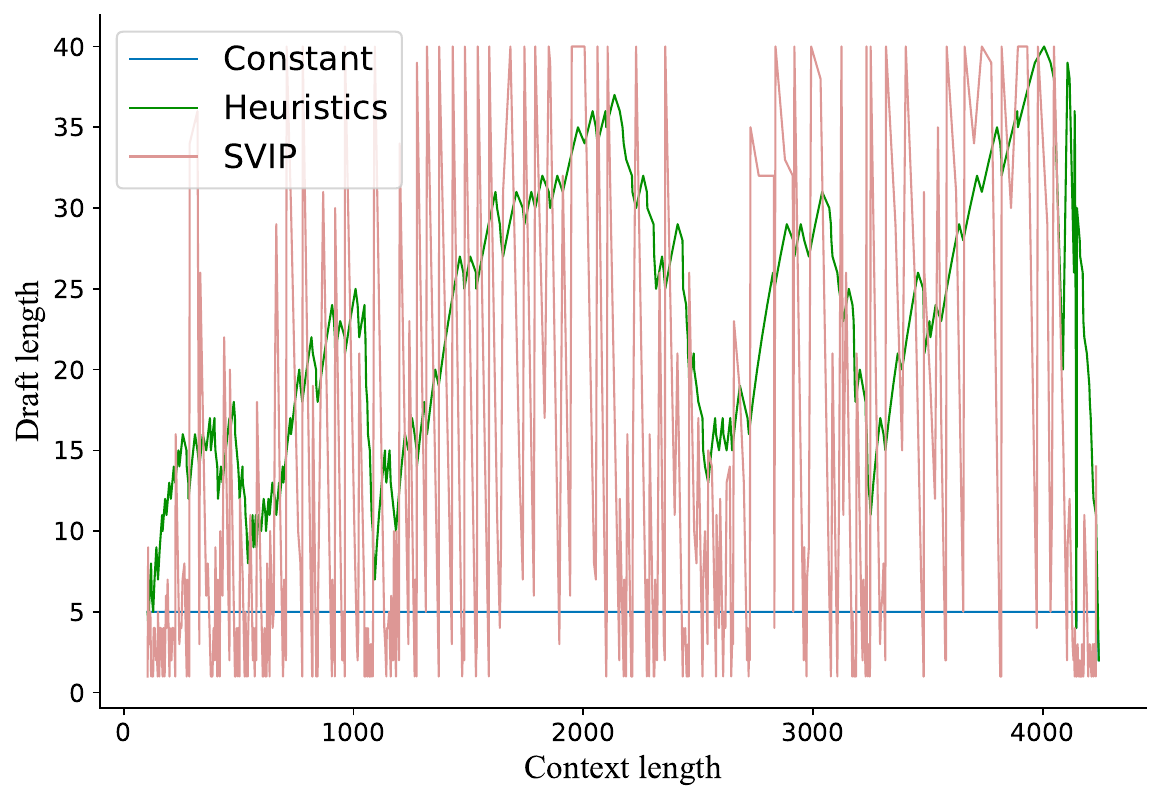}
    \caption{A case comparison of different length policies during one generation example. Drafts proposed by \methodname\ vary drastically in length, while constant and heuristics policies are insufficient to model such changes.}
    \label{fig:qwq-case}
\end{figure}


\begin{table}
\centering
\caption{The average draft model entropy and draft token acceptance rate of some representative tokens between QwQ-32B and 1.5B.}
\label{tab:qwq-tokens}
\resizebox{0.8\linewidth}{!}{%
\begin{tabular}{lrr} 
\toprule
\textbf{Token} & \textbf{Avg. Entropy} & \textbf{Accept Rate} \\ 
\midrule
All tokens & 0.38 & 0.68 \\ 
\midrule
“Wait” & 1.17 & 0.53 \\
“Alright” & 1.38 & 0.22 \\
“Actually” & 1.52 & 0.33 \\
\midrule
“)]” & 0.12 & 1.00 \\
“\}\}” & 0.02 & 1.00 \\
“ynomials” & 0.01 & 1.00 \\
“ponents” & 0.01 & 1.00 \\
\bottomrule
\end{tabular}
}

\end{table}

In Table~\ref{tab:qwq-tokens}, we present some token cases with very low or very high acceptance rate. We find that completing subword units or equations is quite easy for the small draft model, while several keywords in QwQ's reasoning patterns are much harder. The reverse correlation of draft entropy and acceptance rate on these tokens further validates the motivation of \methodname. In Figure~\ref{fig:qwq-case}, we plot the behaviors of different draft length policies in an example generation case. The drastic draft length oscillations in \methodname\ also highlight the importance of dynamic draft length policy.

\section{Related Work}\label{sec:relatedwork}
\vspace{-0.1cm}

Since \citet{2022SD-leviathan} and \citet{2023SD-chen} introduced speculative decoding into large language models, numerous works have followed their tracks in pursuit of more efficient LLM inference. We broadly categorize these works into three types: better draft models, draft tree expansion, and draft length control, which are orthogonal to each other. A more comprehensive review of speculative decoding is provided by \citet{2024Spec-Bench}.

\paragraph{Better draft models.} As \citet{2024Spec-Bench} suggest, draft models in speculative decoding can be either based on self-drafting or based on an independent draft model. For the first type, one may use a quantized~\citep{2024QSpec}, early-exiting~\citep{2024LayerSkip}, or forward-padded~\citep{2023PaSS} version of the target model to produce draft tokens, while the second type is represented by the vanilla speculative decoding~\citep{2022SD-leviathan}. Some works also take the best of both worlds and introduce extra layers on top of the target model's hidden representations to construct draft models, represented by EAGLE~\citep{2024EAGLE}, GliDe~\citep{2024Glide-Cape}, and Medusa~\citep{2024Medusa}. 

Unlike previous methods, \methodname~ has no requirement for draft models except for autoregression, and is a totally training-free adaptive-length policy, which could boost any draft model's performance.

\paragraph{Draft tree expansion.} Given a draft model, one may verify multiple draft tokens for the same position in parallel to increase the probability of finding an accepted draft token, and we use ``draft tree expansion'' as an umbrella term for such techniques. \citet{2024EAGLE-2} introduce EAGLE-2, which reranks draft tokens in EAGLE's draft tree to select tokens with the highest confidence for verification. Similarly, CaPE~\citep{2024Glide-Cape} improves GliDe by expanding the token set chosen for verification at each position based on top-1 confidence. Other works have also addressed the problem of multi-draft verification from a theoretic perspective~\citep{2023SpecTr,2024theoretical-perspective}. 

In contrast to previous methods which introduce tree expansion for proposing fixed fine-grained n-gram draft, our method \methodname\ involves no fixed tree expansion rules, and is mainly about a more dynamic and flexible draft length policy.

\paragraph{Draft length control.} Works in this category are few, but most relevant to ours. \citet{2024PEARL} introduce PEARL, which lets the target model perform verification in parallel to draft generation, stopping the draft process when a mismatch is found. \citet{2024SpecDec++} propose SpecDec++, which trains an acceptance prediction head on top of the draft model to predict the acceptance probability of the current draft token, stopping the draft round when the predicted acceptance probability falls below a constant threshold. \citet{2024DDD} propose Dynamic Depth Decoding (DDD) on top of EAGLE-2, which uses the sum of all tokens' confidences in one level of its draft tree as an indicator to predict whether or not to continue draft generation. Concurrent with our work, \citet{2024AdaEAGLE} propose AdaEAGLE, which utilizes an MLP on top of EAGLE to predict the next round's draft length. 

In contrast to prior length-control strategies that necessitate the training of a length-prediction module, \methodname\ stands out with its training-free nature. This unique characteristic endows it with remarkable flexibility, allowing it to be seamlessly integrated and applied to any autoregressive draft model. 
\section{Conclusion}\label{sec:conclusion}

We propose \methodname, a flexible, training-free, and plug-and-play dynamic draft length policy for speculative decoding systems. Based on a theoretical lower bound of acceptance probability and its empirical approximation, \methodname\ determines whether to continue draft generation or to quit drafting based on the draft model's entropy after the generation of each draft token. 
With extensive experiments spanning various base models, draft methods, test domains, and generation length, we validated the effectiveness of \methodname, sparking new insights on speculative decoding and more efficient large language models. 
For future work, we aim to investigate tighter bounds on the acceptance rate to improve the accuracy of acceptance probability estimates, thereby enabling more efficient draft length adaptation. Additionally, our current analysis may not fully capture context-dependent patterns. A more nuanced investigation into these patterns could further enhance performance.

\section*{Limitations}

While \textsc{\methodname} advances speculative decoding through adaptive draft length control, it has several limitations that offer avenues for future work. 
The acceptance rate bound in \Cref{eq:appro_bound} could be overly conservative. Developing a tighter bound would enhance the accuracy of acceptance probability estimates, enabling more effective draft length adaptation.
Further, our analysis assumes they follow a simplified distribution discrepancy of SD systems. This may not fully capture the nuanced factors contributing to their occurrence, such as context-dependent patterns or model-specific biases. Context-dependent length proxy for SD systems could be the potential research direction.


\bibliography{example_paper}

\appendix

\appendix
\onecolumn

\section{The Complete Speculative Decoding Algorithms}\label{appendix:sd-algorithm}
In Algorithm~\ref{algo:verify-sample} to \ref{algo:sd}, we present the complete algorithms of the vanilla speculative decoding in both the greedy decoding and the sampling scenarios. For the sampling scenario, the \texttt{Verify} and \texttt{Correct} methods in Algorithm~\ref{algo:sd} resolve to Algorithm~\ref{algo:verify-sample} and \ref{algo:correct-sample}. For greedy decoding, they resolve to Algorithm~\ref{algo:verify-greedy} and \ref{algo:correct-greedy}.

\begin{figure*}[ht]
\begin{minipage}[c]{0.45\textwidth}

\begin{algorithm}[H]\small
\caption{Verify (Sampling)}
\label{algo:verify-sample}
\begin{algorithmic}[1]
    \Require target distribution $p(x)$, draft distribution $q(x)$, draft token $x_t$
    \vspace{3pt}
    \State $accept \gets$ False
    \State $r \sim U[0, 1]$
    \If{$r < \frac{p(x_{t})}{q(x_{t})}$}
         \State $accept \gets$ True
    \EndIf
    \Ensure $accept$
\end{algorithmic}
\end{algorithm}

\vspace{-15pt}

\begin{algorithm}[H]\small
\caption{Verify (Greedy)}
\label{algo:verify-greedy}
\begin{algorithmic}[1]
    \Require target distribution $p(x)$, draft distribution $q(x)$, draft token $x_t$
    \vspace{3pt}
    \State $accept \gets$ False
    \If{$\arg\max p(x) == x_t$}
        \State $accept \gets$ True
    \EndIf
    \Ensure $accept$
\end{algorithmic}
\end{algorithm}

\vspace{-15pt}

\begin{algorithm}[H]\small
\caption{Correct (Sampling)}
\label{algo:correct-sample}
\begin{algorithmic}[1]
    \Require target distribution $p(x)$, draft distribution $q(x)$
    \vspace{3pt}
    \State Sample  $\hat x \sim \frac{\max(q(x)-p(x),\,0)}{\sum_i\max(q(x^{i})-p(x^{i}),\,0)}$
    \Ensure $\hat x$
\end{algorithmic}
\end{algorithm}

\end{minipage}
~~
\begin{minipage}[c]{0.51\textwidth}

\begin{algorithm}[H]\small
\caption{Correct (Greedy)}
\label{algo:correct-greedy}
\begin{algorithmic}[1]
    \Require target distribution $p(x)$, draft distribution $q(x)$
    \vspace{3pt}
    \Ensure $\arg\max p(x)$
\end{algorithmic}
\end{algorithm}

\vspace{-15pt}

\begin{algorithm}[H]\small
\caption{Speculative Decoding}
\label{algo:sd}
\begin{algorithmic}[1]
    \Require target model $p$, draft model $q$, input sequence $x_{\leqslant t}$, maximum length $T$, draft length $\gamma$
    \vspace{3pt}
    \State Initialize $n\gets t$
    \While{$n<T$}
        \For{$j = 1$ to $\gamma$}
        \State Sample $x_{n+j} \sim q(x|x_{<{n+j}})$
        \EndFor
        \State Compute $p(x|x_{<{n+j}}),\; j=1, \cdots, \gamma+1$ in parallel
        \State $\tilde n \gets n$
        \For{$j=1$ to $\gamma$}
            \If{Verify$\left(p(x|x_{<n+j}),\, q(x|x_{<n+j}), x_{n+j}\right)$}
                \State $\tilde n \gets \tilde n + 1$
            \Else
                \State $x_{n+j} \gets \text{Correct}\left(p(x|x_{<n+j}),\, q(x|x_{<n+j})\right)$
                \State Exit for loop
            \EndIf
        \EndFor
        \If{$\tilde n == n+\gamma$}
            \State $x_{n+\gamma+1} \sim p(x|x_{\leqslant n+\gamma})$
        \EndIf
        \State $n\gets\tilde n + 1$
    \EndWhile
    \vspace{3pt}
    \Ensure $x_{\leqslant n}$
\end{algorithmic}
\end{algorithm}

\end{minipage}
\end{figure*}

We note that from Algorithm~\ref{algo:verify-sample}, it's straightforward that the acceptance rate of a draft token $x_t$ in the sampling scenario is by definition $\min(1, \frac{p(x_{t})}{q(x_{t})})$.

\section{Alternatives for Acceptance Rate Lower Bound Computation}\label{appendix:lower-bound-alternative}

In Section~\ref{sec:method}, we used Pinsker's inequality to compute a lower bound for the expected acceptance probability:
\begin{align}
    \beta   & = \sum_x \min\left(p(x), q(x)\right) \\
            &\geqslant 1 - \sqrt{\frac{1}{2}\mathbb{KL}(q||p)}.
\end{align}

Another way to compute the lower bound of acceptance probability can be derived from Bretagnolle-Huber inequality~\citep{1978Bretagnolle-Huber}:
\begin{align}
    \beta   &\geqslant 1 - \sqrt{1 - e^{-\mathbb{KL}(q||p)}}.
\end{align}

Compared with the Pinsker's bound, it's trivial to see that this bound is guaranteed to be always larger than 0. However, in practice we find that the Pinsker's bound is about 11\% tighter.

\section{$\gamma$ Approximation}
\subsection{Approximation Bound}
Following Eq.~\eqref{eq:beta-bound}, the acceptance rate $\beta$ satisfies:
$$
    \beta\geq1-\sqrt{\frac{1}{2}\mathbb{KL}(q||p)} = 1-\sqrt{\frac{1}{2}H_{q,p}-\frac{1}{2}H_q}
$$
We denote the above bound as the \textit{actual bound}. While this bound is theoretically sound, it relies on the exact access to the $H_{q,p}$ the cross entropy between target and the draft models, which is inaccessible during the SD drafting phase. To address this, approximate $H_{q,p}$ with $\gamma H_q$, where $\gamma$ is a random variable to describe the ratio between $H_{q,p}$ and $H_q$, i.e. $\gamma = H_{q,p}/H_q$, we could rewrite the bound:
\begin{gather}
    \mathbb{KL}(q||p) = H_{q,p} - H_q = (\gamma-1)H_q \nonumber \\
    \beta\geq 1-\sqrt{\frac{1}{2}(\gamma-1)H_q} \nonumber
\end{gather}
To make this bound more practical, we approximate it using a constant $c$, obtaining
$$
    \beta\geq 1-\sqrt{\frac{1}{2}(\gamma-1)H_q}\approx 1-\sqrt{cH_q}
$$
We denote the above bound as the \textit{approximation bound}.
Since $\gamma$ is a random variable, the tightness and reliability of this approximation depend on how well $c$ aligns with $\gamma$'s behavior. Specifically, we need our approximation bound is smaller than the actual bound:
$$
\beta\geq 1-\sqrt{\frac{1}{2}(\gamma-1)H_q} \geq 1-\sqrt{cH_q}
$$
Simplify the right side inequality:
\begin{gather}
    \gamma \leq 2c+1
\end{gather}

\subsection{Theoretical Analysis}
Now, from the $\gamma$'s distribution in Figure 5, let's analyze the probability that the lower bounds hold by modeling $\gamma$'s distribution. 
\paragraph{Gaussian Distribution}
Let's assume $\gamma\sim N(\mu, \sigma^2)$, and the probability that the bound holds is: 
\begin{equation}
    P(\gamma \leq 2c+1) = \Phi(\frac{2c+1-\mu}{\sigma}) \label{eq:gaussian_assump}
\end{equation}
\noindent where $\Phi$ is the standard normal CDF. 
It demonstrates that:
\begin{itemize}
    \item If $c$ is small (e.g. $2c+1\leq \mu$), the probability that the bound holds is low.
    \item If $c$ is large (e.g. $2c+1\geq \mu$), the bound holds with high probability; however the bound itself becomes loose.
\end{itemize}


\paragraph{Gamma Distribution}
Given the right-skewed nature of $\gamma\geq 1$, we model as a shifted Gamma distribution: $\gamma=1+X$, where $X\sim \text{Gamma}(\alpha,\beta)$. The conditions for the bound to hold is :
\begin{equation}
    P(\gamma\leq 2c+1) = P(X\leq 2c)=\frac{\gamma(\alpha, \beta \cdot 2c)}{\Gamma(\alpha)} \label{eq:gamma_assump-appendix}
\end{equation}

\noindent where $\gamma(\alpha, z)=\int_0^zt^{\alpha-1}e^{-t}dt$ is the lower incomplete gamma function, and $\Gamma(\alpha)$ is the gamma function. This probability depends on $c$, $\alpha$, and $\beta$.
\begin{itemize}
    \item If $c$ is small (e.g. $2c<\mathbb{E}[x]=\alpha/\beta$), so the probability that our approximation further lowers the actual bound is low.
    \item It $c$ is large, the probability approaches 1, but the bound $\beta\geq 1-\sqrt{cH_q}$ is looser.
\end{itemize}

\section{Additional Results on Long-form Generation}\label{appendix:longform-results}




In Table~\ref{tab:long-greedy}, we present the results of greedy decoding using Qwen2.5 14B as target and 0.5B as draft, and find that the speedup ratio of greedy decoding is much higher compared with the sampling experiments in Section~\ref{sec:experiments-mtbench}. Further investigation suggests that this is a result of repetition hallucination in both target and draft models during long-form greedy generation.

\begin{table*}[h!]
    \centering
    \caption{Results of greedy decoding on MT-Bench. Greedy decoding leads to repetition hallucinations in both target and draft models in long-form generation, resulting in exaggerated speedup ratio.}
    \label{tab:long-greedy}
    \begin{tabular}{llcccccccc} 
    \toprule
        \multirow{2}{*}{\textbf{Model}} & \multirow{2}{*}{\textbf{Methods~}}       & \multicolumn{8}{c}{\textbf{Context Length}}    \\ 
        \cmidrule{3-10}
        \multicolumn{2}{l}{}     & 128  & 256  & 512  & 1K   & 2K   & 4K   & 6K   & 8K    \\ 
    \midrule
        \multirow{3}{*}{Qwen2.5 (14B, 0.5B)} & Const. & 1.05 & 1.08 & 1.15 & 1.29 & 1.44 & 1.54 & 1.60 & 1.67 \\
        & Heuristics & 1.04 & 1.06 & 1.13 & 1.32 & 1.54 & 1.72 & 1.85 & 1.97 \\
        & \methodname & \textbf{1.30} & \textbf{1.34} & \textbf{1.42} & \textbf{1.57} & \textbf{1.74} & \textbf{1.87} & \textbf{1.98} & \textbf{2.10} \\
    \bottomrule
    \end{tabular}
\end{table*}


\end{document}